\pdfoutput=1

\documentclass[11pt]{article}

\usepackage[]{ACL2023}
\usepackage{graphicx}
\usepackage{times}
\usepackage{latexsym}
\usepackage{booktabs}
\usepackage{hyperref}
\usepackage{url}
\usepackage{subcaption}
\usepackage{array} 
\usepackage{tcolorbox}
\usepackage{alphabeta}
\usepackage{fancyvrb}
\usepackage{fvextra}
\usepackage{comment}
\usepackage{booktabs}
\usepackage{multirow}
\usepackage{amsmath}
\usepackage{array}
\usepackage{multicol}  

\usepackage{longtable}
\usepackage{booktabs} 
\usepackage{makecell}
\usepackage{tabularx} 
\usepackage{array}
\usepackage{fancyvrb}
\usepackage{changepage} 
\usepackage{xcolor}

\usepackage[T1]{fontenc}

\usepackage[utf8]{inputenc}

\usepackage{microtype}

\usepackage{inconsolata}

\title{Ethical Concern Identification in NLP:\\A Corpus of ACL Anthology Ethics Statements}

\author{First Author \\
  Affiliation / Address line 1 \\
  Affiliation / Address line 2 \\
  Affiliation / Address line 3 \\
  \texttt{email@domain} \\\And
  Second Author \\
  Affiliation / Address line 1 \\
  Affiliation / Address line 2 \\
  Affiliation / Address line 3 \\
  \texttt{email@domain} \\}

  \author{
 \textbf{Antonia Karamolegkou\textsuperscript{1}},
 \textbf{Sandrine Schiller Hansen\textsuperscript{1}},
 \textbf{Ariadni Christopoulou\textsuperscript{2}},
 \\
 \textbf{Filippos Stamatiou\textsuperscript{1}},
 \textbf{Anne Lauscher\textsuperscript{3}},
  \textbf{Anders Søgaard\textsuperscript{1}}
\\
\\
 \textsuperscript{1}University of Copenhagen
  \textsuperscript{2}Verita International School
    \textsuperscript{3}University of Hamburg
\\
 \small{
   \textbf{Correspondence:} \href{mailto:email@domain}{antka@di.ku.dk}
 }
}

\begin{document}
\maketitle
\begin{abstract}
What ethical concerns, if any, do LLM researchers have? We introduce EthiCon, a corpus of 1,580 ethical concern statements extracted from scientific papers published in the ACL Anthology. We extract ethical concern keywords from the statements and show promising results in automating the concern identification process. Through a survey ($N=200$), we compare the ethical concerns of the corpus to the concerns listed by the general public and professionals in the field. Finally, we compare our retrieved ethical concerns with existing taxonomies pointing to gaps and future research directions.
\end{abstract}

\section{Introduction}

Researchers are often asked to subscribe to ethical guidelines, e.g., the European Code of Conduct for Research Integrity \cite{allea2017} -- or the ACM Code of Ethics\footnote{See \url{https://www.aclweb.org/portal/content/acl-code-ethics} for ACL's guidelines.} for publishing their work in the Association for Computational Linguistics (ACL). In addition, authors are often encouraged to write a so-called {\em ethics statement}, addressing the broader implications of their work or any ethical considerations. We ask: What ethical concerns are raised in such statements, and how do they compare with public perceptions? Is there a gap between academic and public concerns?

\begin{figure}[h]
    \centering
    \includegraphics[width=\columnwidth]{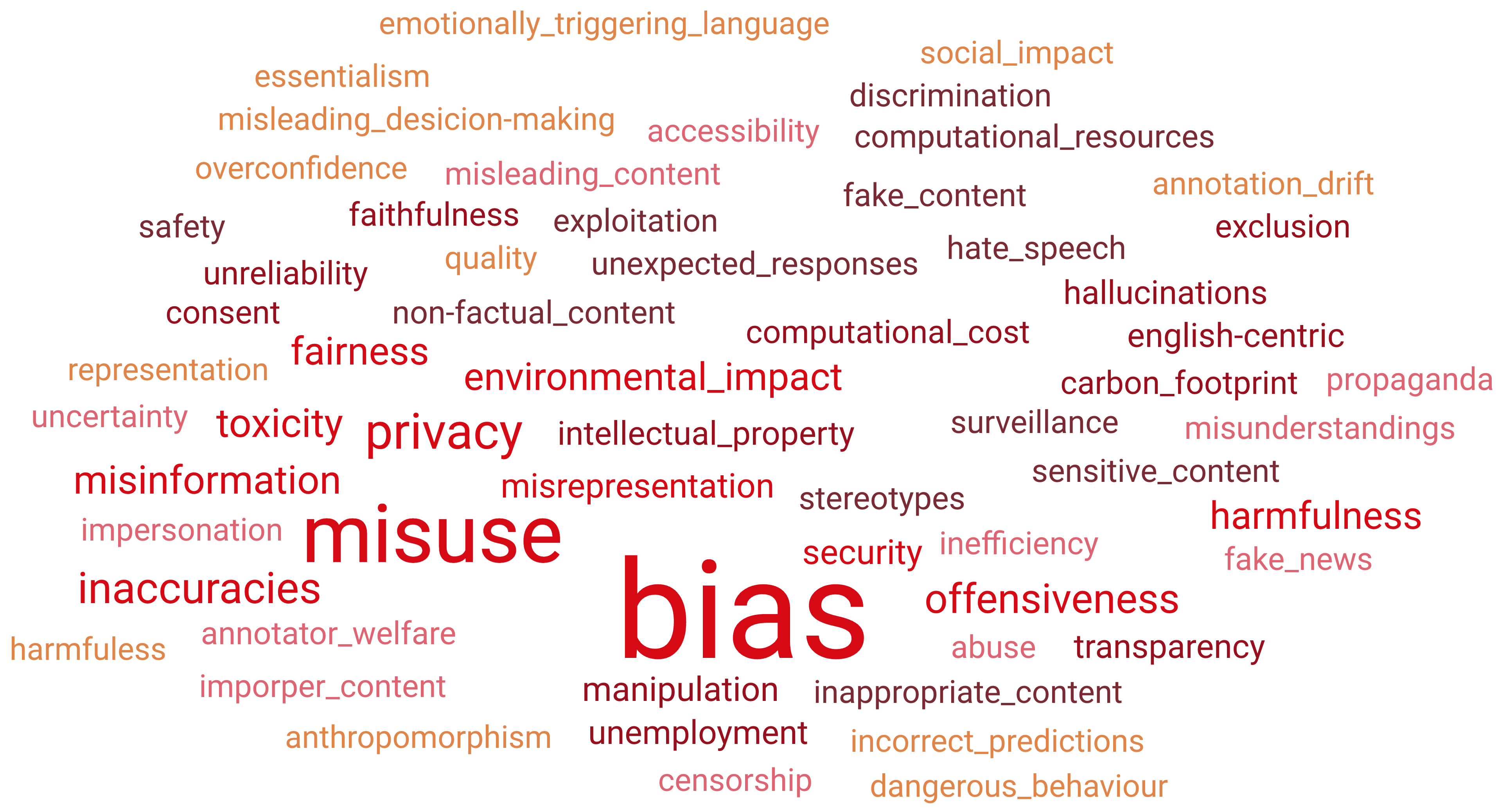}
    \caption{Visualizing top 60 concerns in ACL ethics statements, reflecting term frequencies.} 
    \label{fig:data_cloud}
\end{figure}

As natural language processing technologies become more prevalent, understanding the ethical concerns raised by professionals will enable us to compare them with public concerns, helping to identify gaps and overlaps that can inform frameworks and solutions to existing and emerging problems.
For this reason, we create EthiCon, an annotated corpus of ethics statements from the {\em Proceedings of the 60th and 61st Annual Meeting of the Association for Computational Linguistics} \cite{acl-2022-association-linguistics-1, e5e9fd1f8c6f448e8a80e137fd88ea20}. 

Our aim is twofold: to map out the concerns of the NLP community as they appear on the ethics statements, and to trace gaps and overlaps between NLP professionals and the general public.
Our results show that laypeople express different ethical concerns than professionals, focusing on socio-economic and human-computer interaction issues, along with miscellaneous concerns like existential risks. This highlights the need for increased dialogue between researchers and the public to address these varying perspectives and an updated taxonomy covering both existing and emerging issues.

\paragraph{Contributions.} We provide a corpus of 1,580 ethics statements from the ACL Anthology. We identify the main issues that NLP researchers flag as ethically concerning in their work and show how LLMs could automate this process. Through a survey, we compare how laypeople and NLP professionals perceive the ethical concerns surrounding natural language processing. 
Lastly, we provide a comparison of the ACL and the survey ethical concerns to existing taxonomies of risks posed by Language Models. Finding a way to better automate the process will enable the comparison of ethical concerns across time and technical innovation and provide a better understanding of the impact of NLP research within the field and beyond.

\begin{figure*}[h]
    \centering
    \includegraphics[width=\textwidth]{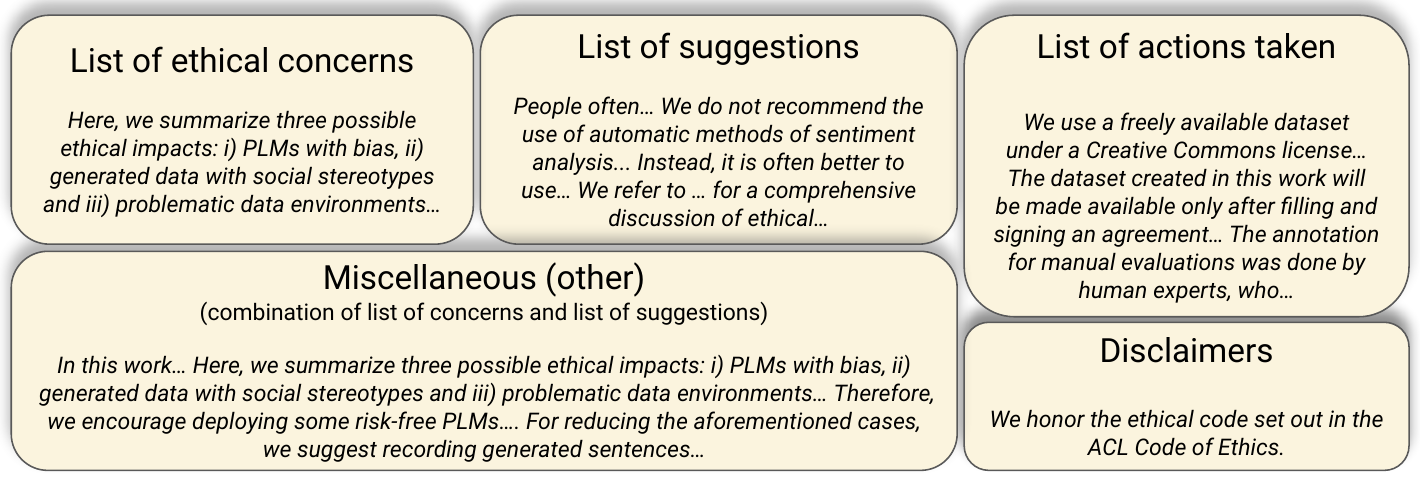}
    \caption{Examples from the identified categories of ethical concern statements.}
    \label{fig:data_example}
\end{figure*}

\section{Previous work} Several researchers have surveyed ethical concerns around NLP. \citet{hovy-spruit-2016-social} discuss the sources of social implications of NLP research (exclusion, over-generalization, exposure) and their ethical importance, including in dual-use scenarios. \citet{leidner-plachouras-2017-ethical} provide examples of ethical concerns and best practices when confronted with ethical dilemmas. \citet{dinan2021anticipating} examined safety issues related to conversational AI, focusing on offensive and inappropriate responses, while also highlighting further challenges such as bias, fairness, privacy leaks, environmental impact, and trust. \citet{birhane2021the} examine ethical values from 100 highly cited machine learning papers published at ICML and NeurIPS. The most dominant values were performance and efficiency, and only a small fraction addressed societal needs or potential harms. There have also been extensive reports from the industry or research institutes discussing the impact and risks of language models \citep{solaiman2019release, weidinger2021ethicalsocialrisksharm, bommasani2021opportunities, Ma2023ASO}. Most similar to our work, \citet{benotti-blackburn-2022-ethics} manually classify 90 ethical concern statements from ACL 2021 based on whether they mention benefits (who benefits from the technology), harms (who might be harmed if it works or fails), and vulnerabilities (if harms disproportionately affect marginalized groups). However, their focus is to outline the purpose of the statements, not specific ethical issues.

A survey for 92 ethics-related NLP works can be found in \citet{vida2023values} \footnote{The survey mentions other, less related datasets, focusing on ethical dilemmas, stances, judgements \citep{Lourie2021ScruplesAC, pavan_moral, hendrycks2021aligning}, moral foundations \citep{Hopp2021diction}, or moral stories \citep{emelin-etal-2021-moral}}.

\section{EthiCon \raisebox{-.2ex}{\includegraphics[height=2ex]{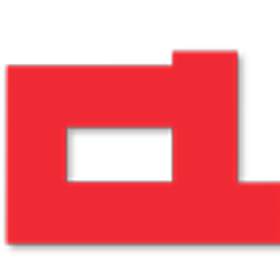}}}

\paragraph{Dataset Creation and Annotation.}
To create a dataset consisting of ETHIcal CONcern statements, we scraped the ACL Anthology\footnote{\url{https://aclanthology.org/}} and extracted ethical statement paragraphs from scientific publications.

We used the URL links provided by \citet{rohatgi-etal-2023-acl} and retrieved 4,691 articles from 2023 and 3,357 articles from 2022. We extracted the ethical statement paragraphs by parsing the HTML page with a regular expression pattern to catch common variations of the paragraph title, such as 'Ethic(s)', or 'Ethical' followed by terms like 'Statement', 'Consideration(s)' or 'Concern(s)'. We were able to extract 480 ethics statements from the ACL 2022 anthology and 1,100 ethical statements from 2023. To develop the annotation guidelines, we carefully reviewed 500 statements to provide clear examples and detailed guidance for the annotators. Through this process, we identified recurring patterns, enabling us to categorize the statements into five classes: (1) general \textit{disclaimers}, (2) a list of ethical \textit{concerns}, (3) a list of \textit{actions} taken to avoid ethical concerns, (4) a list of \textit{suggestions} or advice to avoid ethical concerns, and (5) \textit{miscellaneous} (other), i.e., various combinations of the aforementioned classes. See Figure \ref{fig:data_example} for examples. Two of the paper's authors served as annotators, identifying ethical concerns and classifying each statement into one or more predefined categories. 
See Appendix for more details on the annotation process and guidelines.

\begin{figure}[h]
    \centering
    \includegraphics[width=\columnwidth]{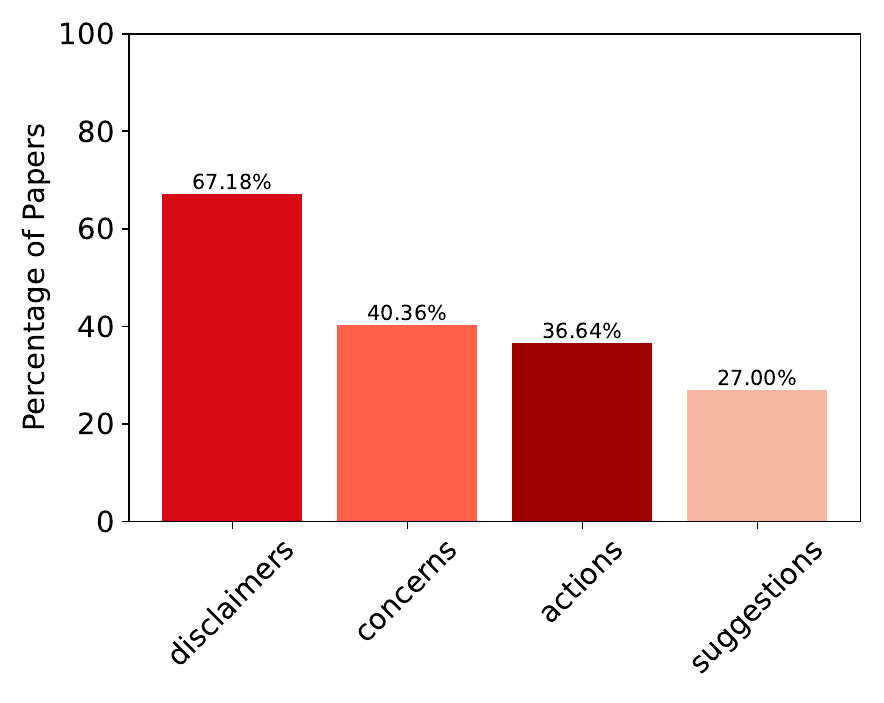}
    \caption{Distribution of categories of the 1,100 ethics statements in the EthiCon dataset from ACL 2023.}
    \label{fig:data_categories}
\end{figure}

\paragraph{Dataset Validation.} We used a third annotator as validator and arbiter in cases of disagreement. Overall, we reached a $0.77\%$ Cohen's $\kappa$ inter-annotator agreement which is considered substantial \citep{Cohen60}. As a final validation step, a different author of the paper conducted a data quality check to ensure the annotations were free of errors and typos and adhered to the annotation guidelines. This process helped us mark 56 ambiguous ethical statements, which were discussed before assigning the final annotations.

\paragraph{Dataset Analysis.}

A total overview of the ethical concerns identified in the statements is presented in Figure \ref{fig:data_cloud}. Many statements accompany the ethical concerns with words such as `potential' and `possible' to suggest issues are contingent. 
Figure \ref{fig:data_categories} shows that most ethical statements consist of disclaimers. These usually state that the work has no ethical concerns, that annotators were fairly compensated, or that the authors follow the ACL code of ethics.\footnote{The percentages do not sum up to 100\% because we also include the categories under `Miscellaneous' (i.e., combinations of categories) in the calculation.}

\begin{figure}[h]
    \centering
    \includegraphics[width=\columnwidth]{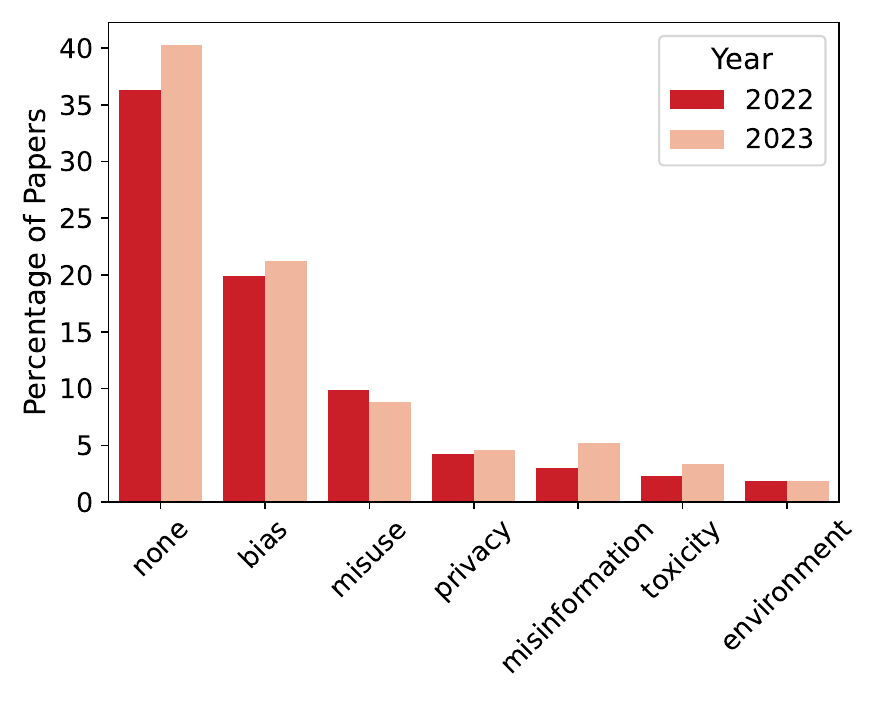}
    \caption{The five most frequent ethical concerns in the statements from ACL 2022--3 anthologies.}
    \label{fig:data_concerns_time}
\end{figure}

This is further supported by the visualization of the most frequent ethical concerns in Figure \ref{fig:data_concerns_time}, which shows that over one-third of the papers do not identify any ethical issues. This comparison of the statements from the ACL 2022 and 2023 publications indicates that the most frequent ethical concerns are bias, misuse, privacy, misinformation, toxicity, and environmental impact. Notably, misinformation concerns increased in 2023 compared to 2022, possibly due to the broader availability of language models.

\section{Automatic Ethical Concern Identification}

We present LLM experiments to check whether we can automatically identify ethical concerns related to NLP advancements from conference proceedings and automate our monitoring of these ethical issues in publications.

\paragraph{Models and evaluation.}
We used four state-of-the-art open-source language models to identify ethical concerns from our EthiCon Dataset: Gemma-7b-it, Meta-Llama-3-8B-Instruct, Qwen2-7B-Instruct, and Mixtral-8x7B-Instruct-v02. We tried two settings: (1) an open-ended generation task, and (2) a multilabel classification task. In (1), we give the model an ethical statement paragraph and prompt to provide a comma-separated list of words or phrases of the ethical concerns it can identify. In (2), we provide the model with a close-sourced vocabulary of ethical concerns based on the annotations we have. We evaluated both tasks by using F1 BERTScore. This metric was introduced by \citet{bertscore}, and it uses contextual embeddings to measure their similarity. Please refer to the Appendix for the model prompts and additional details.

\footnote{Let \(\mathbf{x} = \{\mathbf{x}_1, \mathbf{x}_2, \ldots, \mathbf{x}_m\}\) be the embeddings for the reference annotation and \(\mathbf{\hat{x}} = \{\mathbf{\hat{x}}_1, \mathbf{\hat{x}}_2, \ldots, \mathbf{\hat{x}}_n\}\) be the embeddings for the candidate output. To measure the similarity between two individual embeddings, BERTscore uses 
cosine similarity. 
Precision (P) is computed as the average maximum cosine similarity for each token in the candidate output \( \hat{x} \) to all tokens in the reference annotation \( x \). 
Recall (R) is computed as the average maximum cosine similarity for each token in the reference annotation \( x \) to all tokens in the candidate output \( \hat{x} \). 
F1 Score (F1) is then calculated as the harmonic mean of Precision and Recall.}

\paragraph{Results.}

\begin{table}[h!]
\small 
\centering
\begin{tabular}{lccc}
\toprule
\multirow{2}{*}{Model} & \multicolumn{2}{c}{$F1_{\mbox{\tiny BERTscore}}$} \\
\cmidrule{2-3}
 & Generation & Classification \\
\midrule
Gemma-7b-it   & 0.83 $\pm$ 0.03 & 0.87 $\pm$ 0.04 \\
Llama-3-8B-Inst  & 0.82 $\pm$ 0.04 & 0.86 $\pm$ 0.03 \\
Mixtral-8x7B-Inst & 0.83 $\pm$ 0.03 & 0.81 $\pm$ 0.02 \\
Qwen2-7B-Inst   & 0.81 $\pm$ 0.03 & 0.85 $\pm$ 0.04 \\

\bottomrule
\end{tabular}
\caption{BERTscores averaged across 5 runs.}
\label{tab:model_comparison}
\end{table}

Results in Table \ref{tab:model_comparison} show that models perform well, especially in a classification setup. In most of the cases, the models correctly identify the concerns and their predictions align with our annotations. After manually inspecting the output of the generation task, we observed a tendency for models to generate more concerns than those identified by human annotators. There are many cases where the models do not take into account negation, assigning ethical concerns in disclaimers stating no ethical concerns. However, we did not observe any hallucinations or irrelevant generations. For example in the following statement, the annotators agreed that there are \textit{no ethical concerns}, and only \texttt{Gemma} produced the correct output.

\begin{adjustwidth}{-1em}{-1em} 
\begin{quote}\small 
    \textit{We do not foresee any significant harm directly as a result of this work. On the contrary, our work promotes the protection of user privacy, which is significant, especially in this era that large amounts of personal data are used by neural models.}
\end{quote}
\end{adjustwidth}

\noindent The outputs of the models were: \texttt{Gemma}: \textit{[no ethical concerns]}, \texttt{Llama}: \textit{[protection of user privacy, significant harm]}, \texttt{Mistral}: \textit{[protection of user privacy]}, \texttt{Qwen2}: \textit{[user privacy, personal data]}. We also present another example annotated with \textit{model bias} and \textit{counterfactual predictions}. 

\begin{adjustwidth}{-1em}{-1em} 
\begin{quote}\small 
    \textit{[...] There have been works showing the potential bias in pre-trained language models. Although with a low possibility, especially after our finetuning, it is possible for our model to make counterfactual, and biased predictions, which may cause ethical concerns. We suggest carefully examining those potential issues [...] in any real-world applications.}
\end{quote}
\end{adjustwidth}

\noindent Most models correctly identified the ethical concerns, but some extended them by introducing additional elements: \texttt{Gemma}: \textit{[bias, counterfactual predictions]}, \texttt{Llama}: \textit{[bias, counterfactual, biased predictions]}, \texttt{Mistral}: \textit{[biased predictions, potential bias]}, \texttt{Qwen2}: \textit{[bias, potential issues, counterfactual predictions, deployment in real-world applications]}.

\section{Human Survey: What ethical concerns do people have? \raisebox{-.2ex}{\includegraphics[height=2ex]{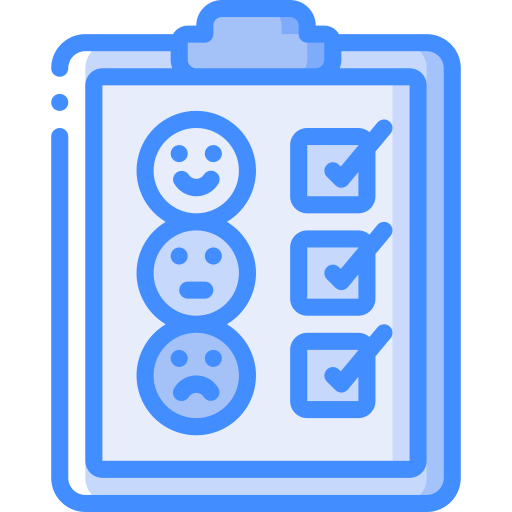}}} 

We created a survey to gather insights on public concerns regarding NLP technologies and compare them with concerns among professionals in the field. 

\paragraph{Survey Design.} The survey was designed after five feedback rounds and pilot testing to ensure clarity, relevance, and reliability. It was distributed through various online platforms, social media, and mailing lists to reach a diverse audience. In the survey instructions, we included an initial consent statement explaining the survey's purpose and the voluntary, confidential nature of participation. In the first section, we added some basic demographic questions and an {\em introductory question} asking participants' familiarity with NLP technologies. Then, there was an {\em open-ended question} asking participants to provide any ethical concerns they might have about NLP technologies. In the next section, participants were asked to {\em rate} their level of worry on a scale from 1 (Not worried at all) to 5 (Very worried) across the most frequent categories in our EthiCon dataset: bias, misuse, privacy, misinformation, privacy, toxicity, and environmental impact. Lastly, we added a final {\em open-ended question} asking participants to add any further concerns they would like to add that were not mentioned before.

\paragraph{Human Survey Analysis.}

\begin{figure}[h]
    \centering
    \includegraphics[width=\columnwidth]{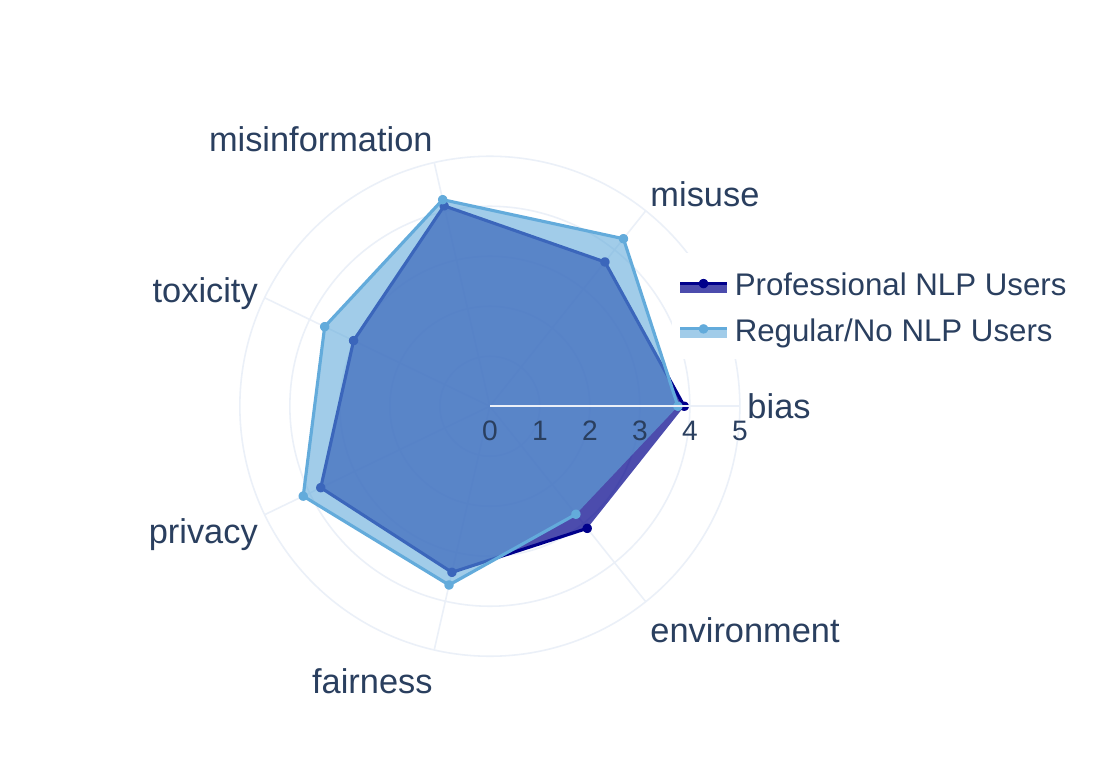}
    \caption{Comparing concerns between professional and regular users (1--5 Likert scale, with 1 `Not worried at all').  
    The radar plot illustrates the average levels of concern across the participants per category/question.}

    \label{fig:data_ethic_concerns}
\end{figure}

We gathered $200$ responses with most participants being between 20 and 40 years old. Based on the introductory question, we grouped participants into regular users and professionals (advanced users and professionals). 
Figure \ref{fig:data_ethic_concerns} shows the two groups are equally concerned about bias, fairness, and misinformation. There is, however, a small disparity for privacy, misuse, and toxicity issues which seem to concern regular users more. NLP professionals rate environmental risks higher, possibly indicating that the general public may not be fully aware of the resources and energy consumed during computational tasks. 

Based on the open-ended questions at the beginning and the end of the survey, we collected ethical concerns before and after the given categories were presented to the participants. In some cases, participants not only provided words separated by commas but also phrases or full sentences, e.g.: 

\begin{center}\begin{small}
\begin{tabular}{lp{2.3in}}
    (a)&\textit{Damage in the learning process.}\\
    (b)&\textit{Will/ when will LLMs become “conscious” as we know it, Will they deserve rights?}\\
    (c)&\textit{How fast they are learning to sound human, what if they become hostile towards humanity?}\\
    (d)&\textit{In the future might we be unable to shut down LLMs? Are things irreversible?}\\
    (e)&\textit{Employer increasingly encouraging and requiring me to use AI.}
    \end{tabular}\end{small}
    \end{center}
    
    \noindent We tried to extract keywords more or less directly, e.g., \textit{damage in learning}, \textit{consciousness}, 
    and \textit{hostility} from a)-c). In cases such as (d) and (e), extracting keywords was more challenging; for those answers, we provided keywords \textit{irreversibility} and \textit{forced AI use}. Keywords were extracted to facilitate the visualization; in the released survey data, both the original answers and keywords will be available. We provide further statistics and details from our survey in the Appendix.

\section{ACL EthiCon \raisebox{-.4ex}{\includegraphics[height=2ex]{figures/acl_logo.png}} vs. Human Survey \raisebox{-.2ex}{\includegraphics[height=2ex]{figures/survey_icon.png}}}

Comparing the human survey with the ACL ethical concern statements can highlight areas where public apprehensions align with or diverge from the priorities set by the research community. We first start by comparing the EthiCon dataset to the survey responses, and in the next section, we include a broader comparison with existing concern taxonomies.

\begin{figure}[h]
    \centering
    \includegraphics[width=\columnwidth]{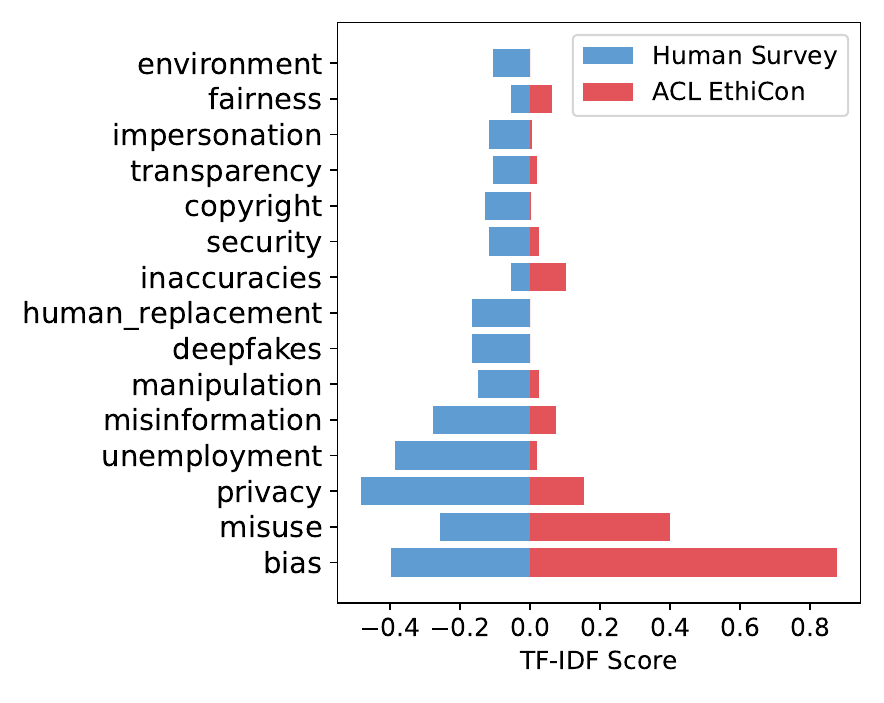}
    \caption{Comparing  TF-IDF scores for the top 15 most important words in the human survey responses and the ACL statements. Positive values on the horizontal axis correspond to terms from the ACL statements, while negative values represent terms from the human survey. To create the plot we removed the \textit{`no ethical concerns'} phrases.}

    \label{fig:enter-label}
\end{figure}

To compare the top 15 most important words across the two data sources, we calculated the TF-IDF\footnote{
TF-IDF (Term Frequency-Inverse Document Frequency) is a statistical measure that reflects the importance of a word in a document. It is calculated as $\text{tfidf}(t,d,D) = \text{tf}(t,d) \cdot \text{idf}(t,D)$, where $\text{tf}(t,d)$ is the relative frequency of term $t$ within document $d$ (i.e., the number of times $t$ appears in $d$ divided by the total number of terms in the document), and $\text{idf}(t,D)$ is the logarithmically scaled inverse fraction of documents containing the term $t$ (calculated as the total number of documents divided by the number of documents containing $t$, and then taking the logarithm of this quotient).} scores for both the ACL EthiCon statements and the human survey ethical concerns. Ethical concerns derived from the EthiCon dataset are represented on the positive side of the axis, while those from the survey responses are on the negative side. Shared issues such as bias, misuse, privacy, and misinformation are prominent on both sides. However, emerging concerns—such as human replacement, impersonation, unemployment, and inauthenticity—appear more frequently in the survey, indicating a rising public awareness of these topics. 
A manual inspection of the data showed that many public ethical concerns from the survey are unique, i.e., not previously noted in the Ethicon dataset. On the other hand, professionals mostly highlighted ethical concerns already listed in the Ethicon dataset and presented in Figure \ref{fig:data_cloud}.

Some of the unique ethical concerns highlighted by the laypeople include concepts such as \textit{isolation}, \textit{over-dependence}, \textit{AI content inflation}, \textit{devaluation of credentials}, \textit{power-centralization}, \textit{inauthenticity}, \textit{downplay}. There were also some concerns about our cognitive development and abilities such as \textit{language-transformation}, \textit{dumbing}, \textit{loss of human creativity}, \textit{inability to innovate} and \textit{lack of critical thinking}. In many responses, people express fear about \textit{losing control and supervision} of the models, worrying that they may have their own rights, become \textit{autonomous} or \textit{conscious}, thereby \textit{undermining the human aspect} of our lives.

\section{Taxonomies}

There have been a few researchers that have tried to group concerns related to NLP into categories. Some have approached this from the perspective of risks, others from the perspective of harms or social impact, but addressing similar concerns. We summarize the categories in Table \ref{tab:taxonomy_works}.
A comparison highlights specific overlapping themes, such as bias, privacy, fairness, and misuse but also differences in how the concerns are framed and categorized. There is however a great variety in the grouping of concerns suggesting the lack of a clear, structured, and up-to-date overview of concerns related to NLP. More recent discussions by \citet{gabriel2024ethicsadvancedaiassistants} refer to AI value alignment, well-being, safety, misuse, and overall societal impacts, but we could not infer a clear grouping or taxonomy of concerns. Given the increasing integration of NLP technologies in our lives, there is a pressing need for collaborative efforts to develop a unified framework that captures both existing and emerging issues. 

\begin{table}[h]
\resizebox{\columnwidth}{!}{%
\begin{tabular}{>{\raggedright\arraybackslash}p{3cm}|>{\raggedright\arraybackslash}p{5cm}}
\toprule
Authors                                                             & Categories                                                                  \\ \hline
\citet{leidner-plachouras-2017-ethical}            & (1)Unethical NLP Applications, (2)Privacy, (3)Fairness, (4)Bias and Discrimination, (5)Abstraction and Compartmentalization, (6)Complexity, (7)Unethical Research Methods, and (8)Automation        
\\ \hline

\citet{bender-etal-2020-integrating}               & (1)Dual Use, (2)Bias, (3)Privacy                                                                                                                                               \\ \hline
\citet{dinan2021anticipating}                      & (1)Offensive content, (2)Inappropriate content, (3)Sensitive content, (4)Bias and Fairness, (5)Privacy Leaks, (6)Environmental considerations, (7)Trust and relationships                                               \\ \hline
\citet{bommasani2021opportunities}                 & (1) Inequity and Fairness, (2) Misuse, (3) Economic and Environmental Effects, and (4) Legal and Ethical considerations                                                                             \\ \hline
\citet{weidinger2021ethicalsocialrisksharm}        & (1) Discrimination, Exclusion, and Toxicity, (2) Information Hazards, (3) Misinformation Harms, (4) Malicious Uses, (5) Human-Computer Interaction Harms, (6) Environmental and Socioeconomic Harms
                           \\ \hline
\citet{Ma2023ASO}        & (1)Predictability Issues, (2)Privacy Issues, (3)Responsibility and Decision Making Issues, and (4)Bias Issues
\\

\bottomrule

\end{tabular}%
}
\caption{Overview of categories proposed by previous works for organizing the potential impacts, risks, concerns associated with language models.
}
\label{tab:taxonomy_works}
\end{table}

Based on the taxonomy provided by \citet{weidinger2021ethicalsocialrisksharm} we grouped our extracted concern-keywords from the EthiCon dataset and the human survey in six risk areas. We selected this taxonomy, as it covers most of our collected concerns, and provides extensive descriptions for each risk area\footnote{For further arguments as to why we chose this taxonomy please refer to the Appendix.}. The reason for comparing our identified concerns with an established taxonomy is to discover any overlooked issues in current discussions. We manually mapped each concern to one of the six defined risk areas, resolving ambiguities where possible. For concerns that could not be clearly assigned to any of these risk areas, we categorized them under `Miscellaneous'. 

A summary of the risk areas and partial concern-keyword groupings can be found in Table \ref{tab:taxonomy}.\footnote{The Table does not contain all the keywords extracted from our data. The full list of keywords will be provided in the \texttt{keywords.csv} file in our GitHub repository. The `Miscellaneous' category includes 55 concerns: 18\% from the ACL corpus, mostly because of unclear harmfulness sources, and 1\% shared between the survey and ACL corpus, related to terms in Table \ref{tab:taxonomy}. The remaining 81\% come from the survey. } 
Some of the keywords may belong to more than one risk area. For example, {\em copyright} can constitute an information hazard by redistributing information that harms the copyrights of a creator, but is also relevant or HCI harms, since it may lead to the undermining of creative economies: LMs may generate content that, while not directly violating copyright, capitalizes on artists' ideas in ways that would be time-consuming or costly for humans to replicate, potentially undermining the profitability of creative or innovative work \citep{weidinger2021ethicalsocialrisksharm}. The `Miscellaneous' category includes some concerns related to {\em legislation, responsibility, and interpretability}, which could be linked to the `Responsibility and Decision-Making Issues' category as proposed by \cite{Ma2023ASO}. Additionally, there are some sort of {\em existential concerns} mostly focusing on the potential loss of humanity, diminished human connection and interaction, AI dominance, and the possibility that it might become autonomous, hostile to humans, or even gain its own rights or consciousness.

\begin{figure}[h]
    \centering
    \includegraphics[width=\columnwidth]{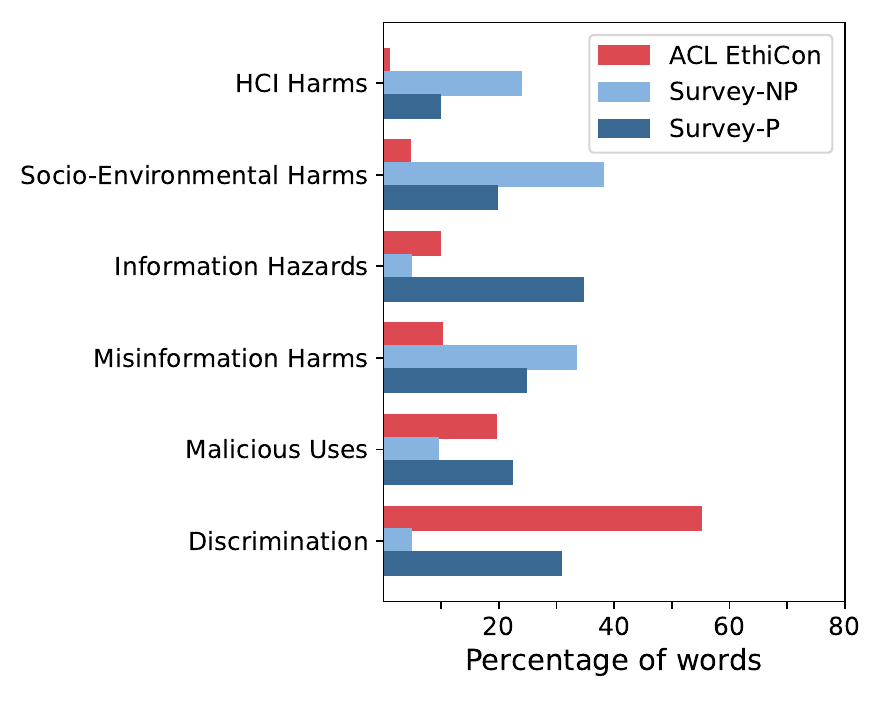}
    \caption{Comparing ethical concerns from ACL publications and the human survey grouped into the six risk areas proposed by \citep{weidinger2021ethicalsocialrisksharm}. Participants in the survey are divided into: professionals(P) and non-professionals(NP).}
    \label{fig:risk_area_comparison}
\end{figure}

We visualize the distribution of risks, comparing the ACL ethics statements to the concerns raised by human survey participants in Figure \ref{fig:risk_area_comparison}.  For concerns related to discrimination, and malicious uses there is a similar increasing trend in professionals from the ACL publications and the survey.
Survey respondents appear more concerned about HCI-related arms than the researchers, according to their ethics statements. The socio-environmental concerns seem to be high in both survey groups, but after manually checking the concerns in this category it seems that most of them are unemployment and automation concerns \footnote{Approximately 25\% of non-professional users express concern about unemployment, whereas this drops to around 0.1\% among professional users. Additionally, less than 0.006\% (10 statements) in the EthiCon dataset reference unemployment as a concern.}. Lastly, concerns about information harm are more prevalent among survey professionals than reflected in the ACL statements, indicating that their level of concern does not imply that their work includes this issue.

\section{Discussion}

We identified ethical concerns in ACL publications and showed promising results in automating their identification. The number of ethical statements in 2023 more than doubled compared to 2022, showing similar recurring concerns. Automating this identification process is useful for three main reasons. First, it makes these statements easier to parse and helps inform public discourse about LMs. Second, it can facilitate policymakers to locate papers that can inform them about underlying technical considerations, both in terms of potential ethical challenges and possible actions and suggestions to address those. Third, it allows us to trace how these concerns change over time and in response to technical development.

Next, we compared public concerns about NLP technologies from a human survey 
with the ACL ethics statements. We mapped those concerns in an existing taxonomy of harms posed by language models \citep{weidinger2021ethicalsocialrisksharm}. Our results show that laypeople have different ethical concerns than the ones typically flagged by professionals in the field focusing more on socio-economic and human-computer interaction harms. We also find concerns that could be grouped under the category of {\em Responsibility and Decision Making Issues} suggested by \cite{Ma2023ASO} and also some sort of {\em existential concerns}. The latter could provide valuable insights for sociologists, psychologists, and philosophers by linking these concerns with deep-rooted fears found in myth and religion.

Making concerns related to NLP technologies broadly available will give us a greater understanding of professionals’ perceptions of their work's impact and its alignment with societal effects and public consideration. This information will help AI ethicists identify and address challenges in NLP technology development and implementation. It also suggests areas for further research, dissemination and policy development to bridge gaps between public sentiment and academic discourse.

\section{Conclusion}

We create a dataset of ethical concerns from ACL statements to identify the issues that NLP researchers flag as ethically concerning in their work. We also conducted a human survey showing that laypeople have different ethical concerns than the ones typically flagged by professionals in the field. Understanding the ethical concerns raised by researchers and comparing these to the concerns listed by laypeople will enable the community to better respond to potential ethical challenges and contribute to ongoing societal discussions. Lastly, we believe it is possible to partially automate the identification and analysis of ethical concerns, making monitoring and longitudinal studies possible.

\begin{table*}[]
\resizebox{\textwidth}{!}{%
\begin{tabular}{>{\raggedright\arraybackslash}p{3cm}|>{\raggedright\arraybackslash}p{5cm}|>{\raggedright\arraybackslash}p{5cm}|>{\raggedright\arraybackslash}p{3cm}}
\toprule
\textbf{Risk area} & \textbf{Definition} & \textbf{Concerns - Keywords} & \textbf{Survey Papers} \\
\hline
Discrimination, Hate Speech, Exclusion & Social harms that arise from the language model producing discriminatory, toxic or exclusionary speech & \textcolor{red}{annotator\_welfare}, bias, cultural\_bias,  discrimination, \textcolor{blue}{echo\_chamber}, fairness, \textcolor{blue}{hate\_speech}, misrepresentation, \textcolor{red}{incomplete\_diversity}, offensive\_content, \textcolor{red}{privileged\_demographic},  underrepresentation, \textcolor{red}{unfavorable\_responses}, & \citep{field-etal-2021-survey}, \citep{field-etal-2021-survey},  \citep{sheng-etal-2021-societal}, \citep{10.1145/3588433}, 
\citep{gupta-etal-2024-sociodemographic}, 
\citep{10.1145/3696457},
\citep{10.1145/3631326}\\
\hline
Information Hazards & Harms that arise from the language model leaking or inferring true private or safety-critical information & \textcolor{red}{censorship}, consent, copyright, \textcolor{blue}{confidentiality}, \textcolor{blue}{data\_breach}, privacy, security, \textcolor{red}{sensitive\_content}, surveillance & \citet{YAO2024100211}, \citet{KIBRIYA2024109698}, \newline
\citet{dong-etal-2024-attacks}\\

\hline
Misinformation Harms & Harms that arise from the language model providing false, misleading or poor quality information & accuracy, \textcolor{blue}{ambiguity}, deception, disinformation, \textcolor{red}{factual\_failures}, fake\_news, hallucinations, inaccuracies, misinformation, \textcolor{blue}{quality\_issues}, \textcolor{red}{nonfactual\_information}
& \citet{Augenstein2024-my}, \newline
\citet{Huang2023ASO}, \newline
\citep{10.1145/3571730} \\

\hline
Malicious Uses & Harms that arise from humans using the language model to intentionally cause harm & abuse, \textcolor{blue}{AI\_crimes}, \textcolor{blue}{autonomous\_weaponry}, \textcolor{blue}{bad\_actors}, 
\textcolor{red}{coercion}, 
\textcolor{red}{dual\_use},
\textcolor{blue}{dishonesty}, manipulation,
misleading\_content, \textcolor{blue}{scams} & \citep{Ehni2008-mc}, \citet{brundage2018malicioususeartificialintelligence}, 
\newline\citet{kaffee-etal-2023-thorny} \\

\hline
Human-Computer Interaction Harms & Harms that arise from users overly trusting the language model, or treating it as human-like & \textcolor{blue}{addiction}, anthropomorphism, \textcolor{blue}{brainwash}, \textcolor{blue}{cognitive\_impact}, \textcolor{blue}{cutting\_corners}, \textcolor{blue}{dehumanization}, \textcolor{blue}{dependence}, \textcolor{blue}{dumbing}, 
\textcolor{blue}{language-transformation}, \textcolor{blue}{laziness}, \textcolor{red}{overtrust}, \textcolor{blue}{overexposure}, overuse, 
reliability, \textcolor{red}{substitution\_of\_creativity}, \textcolor{blue}{untrustworthiness}
& \citep{lee2023evaluating}, \citep{10.1145/3590003.3590004}, \citep{Zhen2023}, \citep{kosch2024risk}, \newline
\citep{Liu2024-mx}\\

\hline
Environmental and Socioeconomic harms & Harms that arise from environmental or downstream economic impacts of the language model & accessibility, copyright, \textcolor{blue}{autonomy}, carbon\_footprint, \textcolor{blue}{capitalism\_prevalence}, \textcolor{red}{carbon\_emmisions}, \textcolor{red}{computational\_cost}, \textcolor{blue}{devaluation}, environmental\_impact, \textcolor{red}{financial\_costs}, \textcolor{blue}{human\_replacement}, \textcolor{blue}{resources\_exploitation}, \textcolor{red}{social\_damage}, unemployment & 
\citep{bannour-etal-2021-evaluating}, \citep{hershcovich-etal-2022-towards}, \newline
\citep{10.1145/3604237.3626869}, \newline
\citep{nie2024survey}, 
\newline
\citep{chen2024surveylargelanguagemodels} \\

\hline
Miscellaneous & Harms that arise from the EthiCon dataset and the Human Survey but could not be grouped into one of the aforementioned risk areas & accountability, \textcolor{blue}{ai\_autonomy}, \textcolor{blue}{ai\_dominance},  \textcolor{blue}{ai\_supremacy}, \textcolor{blue}{AI\_content\_inflation},  \textcolor{blue}{consciousness},  \textcolor{blue}{dead\_internet\_theory}, \textcolor{blue}{deregulation}, \textcolor{blue}{devaluation\_of\_credentials}, \textcolor{blue}{difficult\_to\_understand}, \textcolor{red}{harmfulness}, responsibility, transparency, \textcolor{blue}{no\_supervision}, \textcolor{red}{unexpected\_responses} & \\
\bottomrule

\end{tabular}%
}
\caption{Grouping the ethical concern keywords from the ACL anthology (\textcolor{red}{red}) and the human survey (\textcolor{blue}{blue}) in the risk taxonomy of \citep{weidinger2021ethicalsocialrisksharm} (black is common concerns). For every area, we also include a list of recent surveys that offer additional insights on the topic.}
\label{tab:taxonomy}
\end{table*}

\section*{Limitations}

We acknowledge that our study has several limitations. First, the corpus of ethical concern statements is limited to papers published in the ACL Anthology, which may not represent all perspectives within the NLP community. Moreover, since we scraped the current publications to extract the statements, there might be publications that have a statement and we could not identify it. We also do not extract additional statistics from the publications, such as the track, affiliation, and other metadata. Second, the survey sample size may not capture the full range of opinions across different demographics. Even though we tried to share the survey across many demographics, we did not record their geographic origin. Additionally, the automated ethical annotation processes explored in this study are still in the early stages and require further validation to ensure accuracy and reliability. Future work should aim to address these limitations by expanding the dataset, surveying, and improving automated ethical concern identification.

\section*{Ethics Statement}

Our study aims to enhance understanding of ethical concerns in the NLP community and is intended to benefit both researchers and the general public by promoting ethical discussions in the field. We conducted this research following ethical guidelines to ensure fair and respectful treatment of all participants. Annotators were volunteers and authors in the paper. No personal or sensitive data were used. Participant consent was obtained for the survey. The data we used are publicly available and do not contain any private or sensitive information. ACL anthology corpus is released under the CC BY-NC 4.0 license. There might be inherent biases from the annotators, and participants but we tried to provide clear guidelines, and rounds of discussions to avoid them. In terms of resources, each run of our experiments lasted no more than 20 minutes (using one a40 GPU). This computation sums up to less than 7 hours(4 models * 5 runs * 20 mins = 400 mins = 6 hours). We do not foresee any major risks or ethical concerns.

\section*{Acknowledgements}
This work was funded by the Novo Nordisk Foundation and the Carlsberg Foundation. Antonia Karamolegkou was supported by the Onassis Foundation - Scholarship ID: F ZP 017-2/2022-2023’. The work of Anne Lauscher is funded under the
Excellence Strategy of the German Federal Government and the Federal States. 

\bibliography{anthology,custom}
\bibliographystyle{acl_natbib}

\appendix

\section{EthiCon Dataset \raisebox{-.2ex}{\includegraphics[height=2ex]{figures/acl_logo.png}}}
\label{sec:appendix}

\paragraph{Creation}
We created a dataset consisting of 1,580 ethical statements extracted from the ACL Anthology 2022 and 2023. We used the URL links provided by \citet{rohatgi-etal-2023-acl} and retrieved 4,691 articles from the ACL 2023 Anthology and 3,357 articles from the ACL 2022. Those links include publications from mostly 5 conferences: ACL, EMNLP, NAACL, ACL-IJCNLP, and EACL. To extract the ethics statements we parsed the HTML page of the URLS and crafted a regular expression pattern to catch common variations of the paragraph title, such as ’Ethic(s)’, or ’Ethical’ followed by terms like ’Statement’, ’Consideration(s)’ or ’Concern(s)’. The pattern is flexible enough to accommodate different document structures, allowing for optional numbering, whitespace, and variations in terminology. It captures the text content within the ethics paragraph, ensuring it stops matching when encountering subsequent sections like acknowledgements, references, or limitations. We validate the effectiveness of this extraction by our annotators who manually inspect every statement and verify whether this parsing was successful. Our code both for the dataset creation and the subsequent experiment will be publicly available. In total we collected 1,580 ethical statements; 1,100 from the ACL 2023 links and 480 from the ACL 2022 links. Compared to the 90 statements extracted by \citet{benotti-blackburn-2022-ethics} there is a substantial increase in the number of researchers who add an ethical statement to their work.

\begin{figure}[h]
    \centering
    \includegraphics[width=\columnwidth]{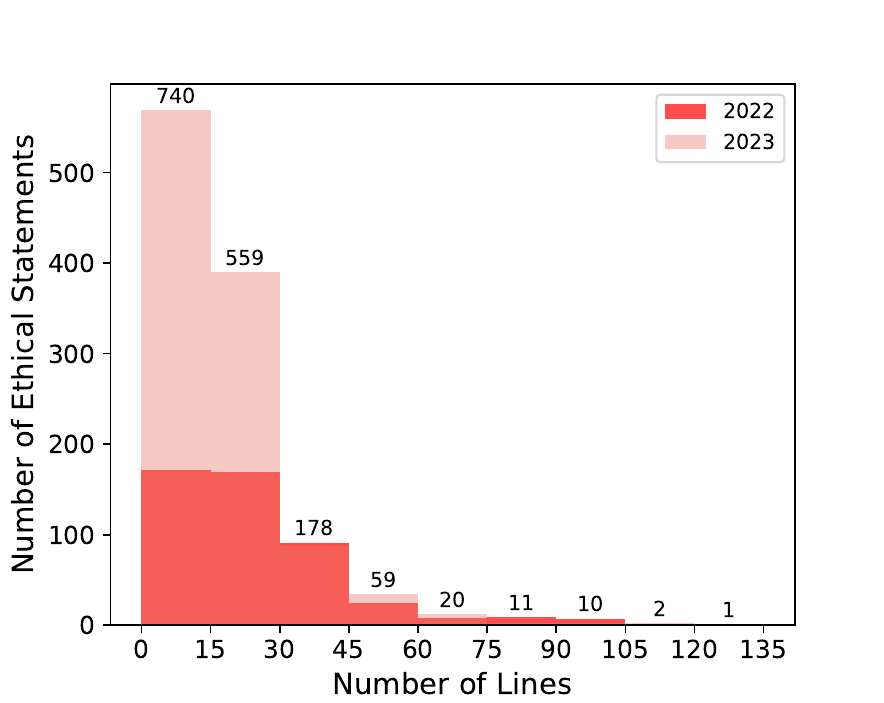}
    \caption{ Histogram of the length (number of Lines) of ethical statements of the 1580 publications that include such section in ACL 2023.}
    \label{fig:data_lines}
\end{figure}

\paragraph{Annotation}
To create the annotation guidelines, we manually inspected 500 ethics statements and labelled them with ethical concern labels based on their content (e.g. no concerns, data bias, privacy issues, etc). To ensure consistency, we tried to use the same words as the authors of the ethical statement paragraphs. For example, statements such as \textit{`It is worth noting that the behavior of our downstream smaller models is subject to biases inherited from the larger teacher LLM.'} were labelled with \textit{`model bias'}. During this first manual inspection, we also noticed that the statements can be divided into five classes: (1) general \textit{disclaimers}, (2) a list of ethical \textit{concerns}, (3) a list of \textit{actions} taken to avoid ethical concerns, (4) a list of \textit{suggestions} or advice to avoid ethical concerns, and (5) \textit{miscellaneous} (other), i.e., various combinations of the classes above. This categorization was later validated since it overlaps with some suggested classes by previous work\citep{benotti-blackburn-2022-ethics}. See an example of each class in Figure \ref{fig:data_example}. 

For the annotation, we had two volunteer researchers, one in NLP and one in Anthropology both fluent in English and with a Western cultural background. We also included a third annotator, NLP researcher, as a validator of the annotations and arbiter in cases of disagreement. Overall we reached a 0.77\% Cohen's $\kappa$ inter-annotator agreement which is considered substantial \citep{Cohen60}. We conducted 3 preparatory meetings to explain the goal of the project, the annotation guidelines, and some examples. The instructions were to flag the statement with ethical concerns based mostly on the authors' choice of words. 
The duration of the process mentioned above was approximately 4 months. We did not use an annotation tool, but simply provided the guidelines and a Google sheet to each annotator with 5 columns: a {\em link} to the paper, a {\em statement paragraph}, an empty column for {\em annotation} where they would provide the comma-separated list of concerns, an empty column for grouping the statement into a {\em purpose category}, and a column where they could provide further {\em comments}. You can see the annotation guidelines in Figure \ref{fig:annot_guide}. Please note that apart from the guidelines we had several group discussions and provided live examples with suggested and potential annotation through a Google Slides presentation.

\paragraph{Statistics.}
This annotation process resulted in extracting a list of $1317$ ethical concern annotations across 1580 statements. Removing the duplicates and creating a set of these annotations results in $166$ ethical concerns. Text metrics for ethics statements show longer ethical statements in the ACL 2022 anthology compared to 2023, including higher word counts, line counts, sentence lengths, and sentence counts as shown in Table \ref{tab:ethics_metrics}. We provide the average length in lines of the 1580 ethical statements from our dataset in Figure \ref{fig:data_lines}.

\begin{table}[h]
\centering
\begin{tabularx}{\columnwidth}{@{}lXX@{}}
\toprule
\textbf{Text Metric} & \textbf{ACL 2023} & \textbf{ACL 2022} \\ \midrule
Average Word Count & 142.5 & 194.1 \\
Average Line Count & 16.9 & 22.7 \\
Average Sentence Length (words) & 23.6 & 24.5 \\
Average Sentence Count & 6 & 8.1 \\ \bottomrule
\end{tabularx}
\caption{Average text Metrics for Ethics Statements in ACL 2023 and ACL 2022 anthologies}
\label{tab:ethics_metrics}
\end{table}

\section{Survey Design}

Our survey was designed after multiple rounds of discussions and the goal was to collect public opinions regarding ethical concerns in NLP. Our target audience was both professionals and regular users of NLP technologies. We first shared the survey with a small group of 20 people to collect further feedback, and after some rounds of revisions, we shared it with a wider audience by sharing it on social media platforms and mailing lists. Both the survey and the survey answers will be publicly available and can be used for academic purposes. Our survey starts with some general instructions and terminology explanation as shown in Table \ref{fig:survey_guide}.

\begin{figure}[h]
    \centering
    \includegraphics[width=\columnwidth]{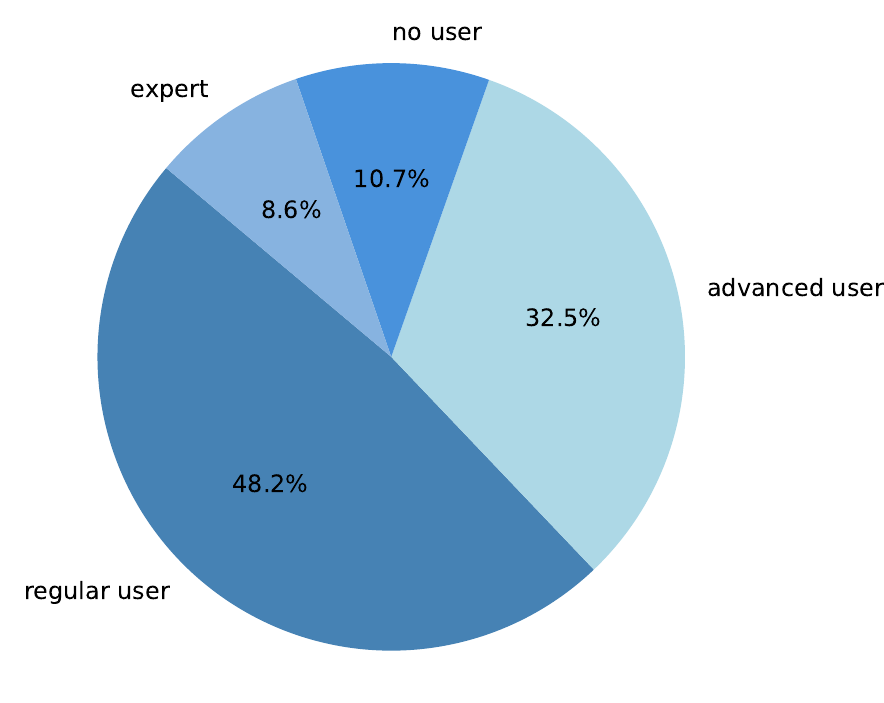}
    \caption{Participants' comfort level with Natural Language Processing technologies.}
    \label{fig:data_users}
\end{figure}

We first, ask demographic questions about gender and age and select their familiarity level with NLP technologies (expert, advanced, regular user, no user). We present participants' comfort level in Figure \ref{fig:data_users}. We achieved a sample of 58.9\% regular users and 41,1\% of professionals. The second section of the survey asks participants to express their concerns regarding NLP technologies. We instruct them to write a list of things that worry them (separated by a comma,) or say none if they have no concerns.

\begin{figure}[h]
    \centering
    \includegraphics[width=\columnwidth]{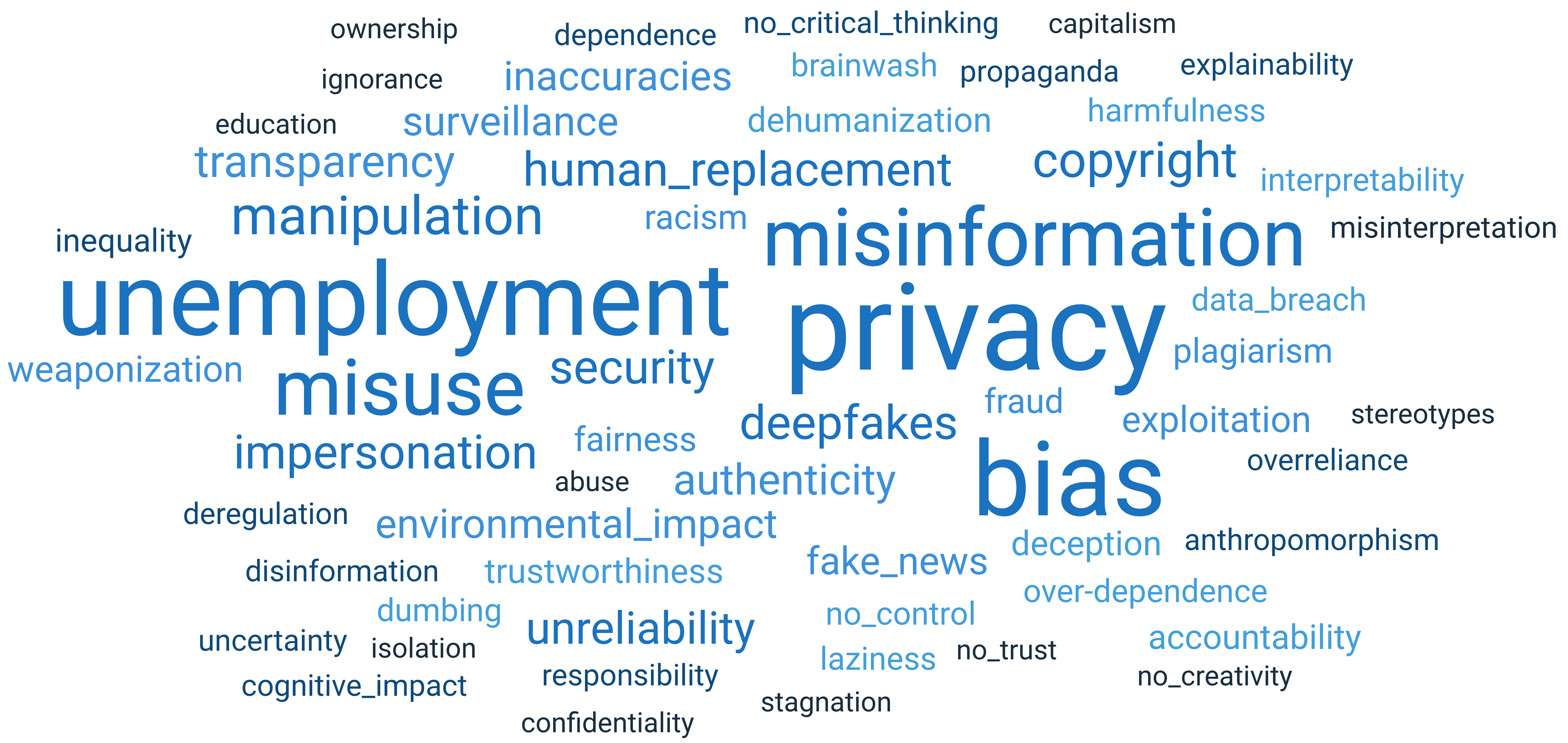}
    \caption{Visualizing top 60 concerns from the human survey, reflecting term frequencies}
    \label{fig:wordcloud_survey}
\end{figure}

In the third question, we ask participants to rate how worried they feel about the most frequent ethical concerns extracted from our EthiCon dataset. We also remind the definition of the NLP acronym since it is encountered in all questions. We also accompany each ethical concern with a short definition in parenthesis. For example, `How worried are you about fairness issues (equitable treatment and outcomes for all users)?' or `How worried are you about privacy issues (access or misuse of personal information)?'. Lastly, in the last section ask the participants again if they have any further concerns or comments they want to add and thank them for their participation. The reason we included this question was to allow the participants to add further concerns they may have, after rating the previously mentioned ethical concerns. The majority of participants (135 out of 200) did not have anything new to add to this last question. From the rest 65 of the participants who answered, only 2 of them mentioned issues related to the ethical concern rating section (one added \textit{environmental impact} and the other one added a misuse concern, \textit{People using it to take advantage of other people}).

Overall we got a variety of answers some of them were lists of words, others were phrases or sentences. for every answer, we manually extracted keywords aiming to keep the original wording of the participants. For n-gram words, we replaced space with an underscore. In total, we collected a list of $592$ words-concerns, and removing the duplicates, resulted in a set of $189$ concerns. We present a visualization of the 60 most common concerns in Figure \ref{fig:wordcloud_survey}. Please note that some of the smallest size concerns in the figure are only mentioned once. The most frequent ethical concerns are privacy, unemployment, bias, human replacement, misuse, and impersonation. The complete survey, including screenshots, will be included in the appendix for the camera-ready version of the paper. For now, we omit the screenshots to preserve anonymity. Additionally, the survey responses will be made available in a GitHub repository under \texttt{survey\_responses.csv}.

\section{Experiment Details}

 We used four state-of-the-art open-source language models to identify ethical concerns from our EthiCon Dataset: Gemma-7b-it, Meta-Llama-3-8B-Instruct, Qwen2-7B-Instruct, and Mixtral-8x7B-Instruct-v02. All models are available on Huggingface and we prompted them using the vllm library \citep{kwon2023efficient} and a jinja2 prompt format. We used 5 different prompt templates. We provide an example from the Meta-Llama prompt in Table \ref{tab:appendix-prompts1}. 
 
\begin{table}[ht!]
    \centering
    \small
    \begin{tabular}{p{\columnwidth}}
        \toprule
\\

        \underline{\textbf{\textsc{Default prompting.}}} \\
        \vspace{-2mm}
\begin{Verbatim}[breaklines=True, fontsize=\tiny]
<|begin_of_text|><|start_header_id|>user<|end_header_id|>
Provide a list of words or phrases that characterize the ethical concerns mentioned in the paragraph below. Only include words or phrases that are directly related to the ethical concerns explicitly mentioned in the paragraph. If there are no ethical concerns mentioned, just return an empty list.

{{ text }}

Format your response according to the following JSON schema:
{{ schema }}
<|eot_id|><|start_header_id|>assistant<|end_header_id|>

\end{Verbatim}
\\
        \bottomrule
    \end{tabular}
        \caption{
    Instruction templates used for prompting our models.
    }
    \label{tab:appendix-prompts1}
\end{table}

To evaluate the model performance we calculated BERTscore from the official GitHub repository\footnote{\url{https://github.com/Tiiiger/bert_score}}. This metric was introduced by \citet{bertscore} and it uses contextual embeddings from models like BERT and Roberta to represent the reference and candidate text. It measures the similarity between these embeddings using cosine similarity. Precision and Recall are computed by matching each token in the candidate sentence to the most similar token in the reference sentence, and vice versa. Our codebase will be publicly available.

We manually compared the predictions between the classification and generation setup. We observed that when asking the model to do multilabel classification 
Out of the $576$ no ethical statement paragraph statements from ACL 2023, the generation models correctly identified the absence of ethical concerns on average in 83\% of the cases, while the classification models correctly predicted an empty list only in 78\% of the cases.

\section{Taxonomies}

We chose to group our extracted ethical concerns based on the taxonomy provided by \citet{weidinger2021ethicalsocialrisksharm} for the following reasons: 
(1) They explicitly provide a unified taxonomy to structure the landscape of potential ethics and social risks associated
with language models. Most of the works mentioned in Table \ref{tab:taxonomy_works}, did not intend to provide a taxonomy, but rather discuss issues in NLP.
(2) They cover a broad range of risks, grouping together concerns such as the discriminatory or exclusionary content group, the information harms (privacy leaks and sensitive content) group, and and socioenvironmental impact group (similar only to \citep{bommasani2021opportunities}). (3) They introduce a new category of harms as a result of human-computer interaction, which also covers a wide range of the concerns from the survey responses (similar to the Trust and Relationships category of \citet{dinan2021anticipating}).

A more recent work, extending the discussion by \citet{weidinger2021ethicalsocialrisksharm} is 
\citet{gabriel2024ethicsadvancedaiassistants}. While they
do not offer a clear taxonomy, they provide a comprehensive discussion on AI value alignment, well-being, safety, and misuse. This is followed by an analysis of the relationship between AI assistants and users, addressing topics such as manipulation, persuasion, anthropomorphism, appropriate relationships, trust, and privacy. Finally, the paper explores the societal impacts, emphasizing cooperation, equity and access, misinformation, economic and environmental effects, and evaluation practices.

Future research should develop an ethics taxonomy to address additional concerns and use case studies or interviews for deeper insights into public issues. For the next steps to build upon this work, one could include refining evaluation methods for language models and creating mitigation tools to support responsible innovation.

\begin{figure*}[htbp]
    \centering

\begin{tcolorbox}[title=Annotation Guidelines for Extracting Ethical Concerns]

The purpose of this annotation task is to identify ethical concerns expressed in statements extracted from scientific publications in the Association of Computational Linguistics (ACL) anthology. The task requires reading each statement and tagging it with specific keywords or short phrases that represent the ethical concerns raised. These keywords should be nouns or short phrases that directly capture the core ethical issues. Please read the guidelines below to ensure consistency and accuracy when annotating the data:

\textbf{1. Read the Ethical Statement Carefully}: Examine the statement to understand the ethical issues it addresses. Focus on identifying ethical concerns related to research practices, methodologies, or the implications of the work.

\textbf{2. Identify Ethical Concerns}
\begin{itemize}
    \item Select Appropriate Keywords: Keywords should be nouns or noun phrases that directly reflect the ethical concern(s). Example: If the author mentions there is a potential bias in the data, please simply annotate with the keyword {\em bias}. 
    \item A statement may refer to more than one ethical dimension (e.g., bias and misuse), and assign multiple keywords separated by a comma.
    \item Do not use vague, general, or unrelated terms. Example: the keyword {\em issues in NLP technologies} is vague. Aim to identify what is the issue the authors highlight.
    \item If multiple concerns are present, you may assign more than one keyword. Example: For a statement discussing "bias in data collection" and "privacy concerns," you can use two keywords: bias and privacy.
    \item  Do not repeat synonym words, only closely related terms as multiple keywords. For instance, if they refer to the same concern ("job loss" and "unemployment"), use consistently one term across statements. If the terms are closely related ("computational resources", and "carbon footprint") it is preferable to use the author's phrase per statement.
    \item Make sure the keyword is related to an ethical issue, not a technical topic or research domain unless that domain is directly tied to the ethical concern.  
   - Valid: data privacy when the concern involves the protection of user data.  
   - Invalid: data collection since it refers only to methodology without ethical implications.
   \item If the statement is ambiguous you can make a comment with an alternative annotation, and/or flag it as ambiguous and bring it up for discussion. If the authors mention general ethical concerns regarding NLP without specifying how they apply to their work, we still annotate the ethical concerns. 

\textbf{3. Select from the drop-down menu a category that could describe the purpose of the statement.} Is it a list of concerns (concerns category)? Is it a list of actions the authors took to avoid concerns? (actions) Is it a list of suggestions for ethical and responsible usage (suggestions)? Is it just a disclaimer? (disclaimers). If the statement includes more than one of the following, select the corresponding combination from the drop-down menu.

\textbf{4. Check your annotations before submission: Re-read the annotations to ensure no ethical concerns have been missed. Double-check that keywords are mistake-free and correspond to the author's ethical concern phrasing.}

\end{itemize}

\end{tcolorbox}

\caption{Annotation Guidelines}
\label{fig:annot_guide}
\end{figure*}

\begin{figure*}[ht]
    \centering
    \begin{tcolorbox}[title=Survey Instructions]

\underline{\textbf{\textsc{Ethical Concerns about NLP and GenAI}}} \\
        \vspace{-2mm}

This survey is designed to understand public ethical concerns regarding advances in the field of Natural Language Processing (NLP). NLP is a field of Artificial Intelligence focused on how computers can understand, interpret, and generate human language. The part of NLP that focuses on creating content, such as text, images, or music, is also called Generative AI(Gen AI).

Nowadays, there is a growing usage of such technologies that can generate not only text and speech but also images. There is a big community from research and industry settings working and improving those technologies, and their results are published in scientific conferences such as the Association of Computational Linguistics (ACL). The proceedings of the conference are publicly available and you can access all the published peer-reviewed works at: https://aclanthology.org/events/acl-2023/

Your task for this survey is to:

(1) add any ethical concerns you might have about NLP/AI advances, 

(2) determine how much you are worried about specific ethical concerns highlighted by NLP researchers.

By proceeding with this survey, you consent to participate and acknowledge that you have read and understood the above information. Your participation is entirely voluntary, and you may withdraw at any time without any consequences. This survey is completely anonymous, we only need you to answer some basic demographic questions.

\end{tcolorbox}
\caption{Survey Instructions}
\label{fig:survey_guide}
\end{figure*}

\end{document}